\documentclass[letterpaper, 10 pt, conference]{ieeeconf}  

\IEEEoverridecommandlockouts                              

\usepackage{floatrow}
\newfloatcommand{capbtabbox}{table}[][\FBwidth]

\usepackage{amssymb}
\usepackage{amsmath}
\usepackage{amsmath,mathtools}
\usepackage{times}
\usepackage{caption}
\usepackage[labelformat=simple]{subcaption}
\usepackage{graphicx}
\usepackage{graphics}
\usepackage{blkarray}
\usepackage{booktabs}
\usepackage{xcolor} 
\usepackage{bbold}
\usepackage{caption}
\usepackage{subcaption}
\usepackage{multirow}
\usepackage{comment}
\usepackage{epstopdf}
\usepackage[section]{placeins}

\usepackage[colorinlistoftodos]{todonotes}

\usepackage{tikz}
\usepackage{todonotes}
\graphicspath{{./figures/}}
\usepackage{csquotes}
\usepackage{siunitx}
\usepackage{cite}
\usepackage{epigraph}
\usepackage{empheq}
\usepackage{comment}
\usepackage{cancel}
\usepackage{algorithm} 
\usepackage{algorithmic}  
\usepackage{mathtools}

\usepackage{balance}
\usepackage{dirtytalk}
\usepackage{pifont}

\usepackage[ruled,vlined,algo2e,noend]{algorithm2e}

\SetInd{0.35em}{0.5em}

\usepackage{hyperref}

\newcommand{\cmark}{\ding{51}}%
\newcommand{\xmark}{\ding{55}}%

\newcommand{\blue}[1]{{\color{black}#1}}
\SetKwRepeat{Do}{do}{while}

    \newcommand{\BTNode}[1]{%
    \begin{tikzpicture}
 
        \node at (0,0) [rectangle,draw, minimum height = 1em] (c) {\tiny #1};
    \end{tikzpicture}
    }

\makeatother

\newdimen\figrasterwd
\figrasterwd\textwidth

\DeclareCaptionLabelSeparator{periodspace}{.\quad}
\makeatletter
\newcommand*{\rom}[1]{\expandafter\@slowromancap\romannumeral #1@}
\makeatother

\def\eqalignno#1{\let\\=\cr\displ@y \tabskip\@centering
  \halign to\displaywidth{\hfil$\@lign\displaystyle{##}$\tabskip\z@skip
    &$\@lign\displaystyle{{}##}$\hfil\tabskip\@centering
    &\llap{$\@lign##$}\tabskip\z@skip\crcr
    #1\crcr}}
\def\leqalignno#1{\let\\=\cr\displ@y \tabskip\@centering
  \halign to\displaywidth{\hfil$\@lign\displaystyle{##}$\tabskip\z@skip
    &$\@lign\displaystyle{{}##}$\hfil\tabskip\@centering
    &\kern-\displaywidth\rlap{$\@lign##$}\tabskip\displaywidth\crcr
    #1\crcr}}
\begin{document}

\pagestyle{empty}
\title{\LARGE \bf
  On the Implementation of Behavior Trees in Robotics}
\author{Michele Colledanchise and Lorenzo Natale  
\thanks{The authors are with the Humanoids Sensing and Perception Lab, Istituto Italiano di Tecnologia. Genoa, Italy.
e-mail: \tt{ michele.colledanchise@iit.it}}}

\maketitle
\thispagestyle{empty}
\pagestyle{empty}

\begin{abstract}
\blue{
There is a growing interest in Behavior Trees (BTs) as a tool to describe and implement robot behaviors. BTs were devised in the video game industry and their adoption in robotics resulted in the development of ad-hoc libraries to design and execute BTs that fit complex robotics software architectures.

While there is broad consensus on how BTs  work, some characteristics rely on the implementation choices done in the specific software library used.

In this letter, we outline practical aspects in the adoption of BTs and the solutions devised by the robotics community to fully exploit the advantages of BTs in real robots. We also overview the solutions proposed in open-source libraries used in robotics, we show how BTs fit in a robotic software architecture, and we present a use case example.
}
\end{abstract}
\section{Introduction}
\label{sec:introduction}


BTs are a graphical mathematical model for reactive and fault-tolerant task executions. 
They were first introduced in the computer game industry \cite{isla2005handling} to control non\--player characters (NPCs)
and is now an established tool that appears in textbooks \cite{millington2009artificial,rabin2014gameAiPro,BTBook} and generic game-coding software such as Pygame, Craft AI, and Unreal Engine. 
BTs are appreciated because they are highly modular, flexible, and reusable. 
 BTs are either created by human experts~\cite{rovida2017extended, guerin2015manufacturing,klockner2015behavior,coronado2018development,paxton2017costar,shepherd2018engineering} or automatically designed using algorithms \cite{pena2012learning,safronov2020task,iovino2020learning} maximizing given objective functions.
The use of BTs in robotics spans from manipulation~\cite{rovida2017extended, zhang2019ikbt,csiszar2017behavior, tenorth2019controlling}
to non-expert programming~\cite{coronado2018development,paxton2017costar,shepherd2018engineering}. Other works include task planning \cite{neufeld2018hybrid}, human-robot interaction~\cite{kim2018architecture, axelsson2019modelling,ghadirzadeh2020human}, learning~\cite{sprague2018adding, banerjee2018autonomous,scheidelearning}, multi-robot systems~\cite{biggar2020framework,tadewos2019fly,kuckling2018behavior,ozkahraman2020combining}, and system analysis~\cite{de2020reconfigurable,ogren2020convergence}.
The Boston Dynamics' Spot SDK uses BTs to model the robot's mission~\cite{spot}; the Navigation Stack of ROS2 and the one of NVIDIA SDK encode robot behaviors as BTs~\cite{macenski2020marathon2}.

Despite the similarity in the deliberation capabilities of NPCs and robots, NPCs are virtual agents that act in a virtual, usually known, environment; whereas robots are physical agents that act in an unknown environment. This required robotics researchers and developers to modify BTs to accommodate the robotic applications' requirements and better integrate them into existing robot software architectures.

\begin{figure}[t]
\centering
\includegraphics[width=0.8\columnwidth]{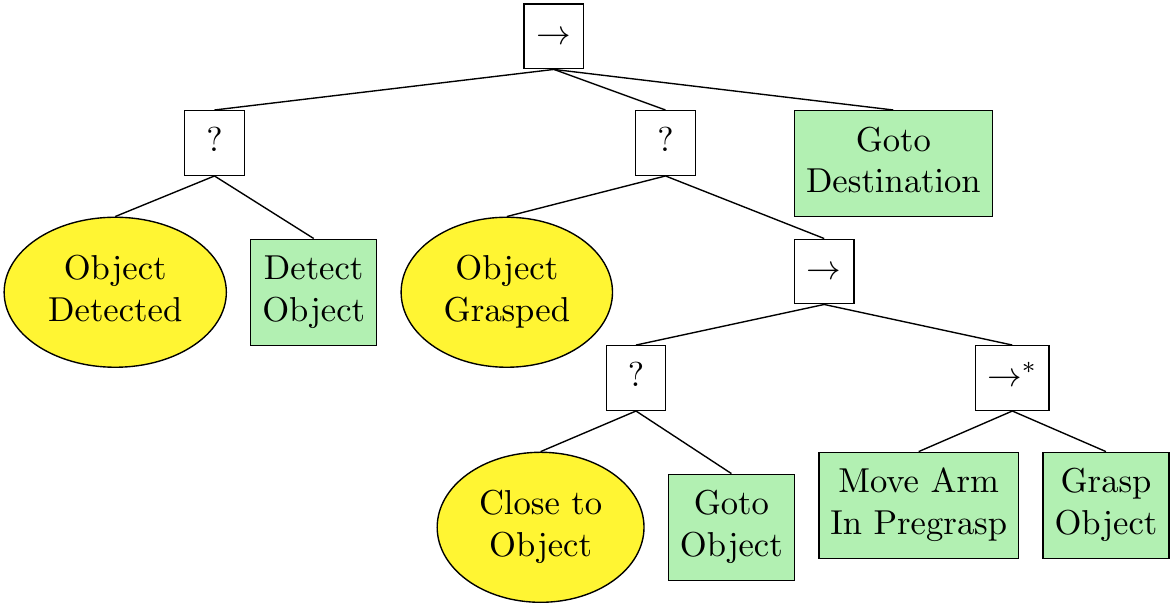}
\caption{BT for a fetch task.}
\label{intro.BT}

\end{figure}

Consider the BT shown in Figure~\ref{intro.BT}. The BT encodes the robot's behavior that has to find an object, approach it, grasp it, and then go to a destination. The implementation and execution of this BT assume the following:
\begin{itemize}
\item Node Templates: The BT has the actions \emph{Goto Object} and \emph{Goto Destination}. This implies the existence of an action template \emph{Goto $<$argument$>$} that can be refined by setting the value of the argument. 
\item Communication mechanism between nodes: the action \say{Detect Object}
computes the 6D pose of the object, such pose gets then used by other nodes like \say{Goto Object} and \say{Grasp Object}. 
\item Asynchronous action execution: The BT activates actions by sending \emph{ticks} to them, as we describe below. The action  \say{Goto Object}, in turn, controls the robot by sending commands to navigate to a given location. The two control loops, the one of the BT's tick traversal and the one of the action execution, may run asynchronously.  
\item Preemption: To support reactivity and safety, the BT may require the running nodes' interruption.  
\end{itemize}

Hence, the simple BT above requires a design and execution engine that handles: parametrization, infra-node communication, asynchronous execution, and safe interruptions.  

\blue{
To the best of our knowledge, while there exists surveys on BTs in robotics and a comparison of software libraries~\cite{iovino2020survey,ghzouli2020behavior}, the literature lacks a study on the implementation details of BTs in robotics. Our intention is to help readers understand the implementation aspects of BTs with their implications in the execution of robot behaviors and suggest practical solutions to integrate BTs in a robotics software architecture.

In this letter, we outline the current solutions adopted by the robotics community. After describing the background on BTs (Section~\ref{sec:bts}), we focus our attention on composition nodes and their stateful counterpart (Section~\ref{sec:memory}), node parametrization and message passing (Section~\ref{sec:param}), asynchronous action execution and the importance of preemptable actions (Section~\ref{sec:syncvsasyc}). We then outline how the most popular open-source libraries reflect these characteristics (Section~\ref{sec:libraries}), we show how BTs fit a robotic architecture (Section~\ref{sec:example}), and finally we present a systematic example (Section~\ref{sec:use}).}

\section{Behavior Trees}

\label{sec:bts}

In this section, we present the classical formulation of BTs. A detailed description of BTs is available in the literature~\cite{BTBook}.

\subsection{Classical Formulation of Behavior Trees}
\label{sec:background.BT}

A BT is a directed rooted tree where the internal nodes represent behavior compositions and leaf nodes represent actuation or sensing operations. We follow the canonical nomenclature for root, parent, and child nodes.

The children of a BT node are placed below it and they are executed  from left to right. The execution of a BT begins from the root node. It sends \emph{ticks}, which are activation signals, with a given frequency to its children. A node in the tree is executed if and only if it receives ticks. When the node no longer receives ticks, its execution stops.  The child returns to the parent a status that can be either \emph{Success}, \emph{Running}, or \emph{Failure} according to the logic of the node. Below we present the most common BT nodes and their logic.

In the classical representation, there are four composition nodes (Fallback, Sequence, Parallel, and Decorator) and two execution nodes (Action and Condition).

\subsubsection*{Sequence}

When a Sequence node receives ticks, it routes them to its children from left to right. It returns Failure or Running whenever a child returns respectively Failure or Running. It returns Success whenever all the children return Success. When child $i$ returns Running or Failure, the Sequence node stops sending ticks to the next child (if any) but keeps ticking all the children up to child $i$.
\blue{
The Sequence node is graphically represented as \BTNode{$\rightarrow$}, }and its pseudocode is described in Algorithm~\ref{bts:alg:sequence}.

%

\begin{algorithm2e}[h!]
\SetKwProg{Fn}{Function}{}{}

\Fn{Tick()}
{
  \For{$i \gets 1$ \KwSty{to} $N$}
  {
    \ArgSty{childStatus} $\gets$ \ArgSty{child($i$)}.\FuncSty{Tick()}\\
    \uIf{\ArgSty{childStatus} $=$ \ArgSty{Running}}
    {
      \Return{Running}
    }
    \ElseIf{\ArgSty{childStatus} $=$ \ArgSty{Failure}}
    {
      \Return{Failure}
    }
  }
  \Return{Success}
  }
  \caption{Pseudocode of a Sequence operator with $N$ children}
  \label{bts:alg:sequence}
\end{algorithm2e}

\subsubsection*{Fallback}
 
When a Fallback node receives ticks, it routes them to its children from left to right. It returns a status of Success or Running whenever a child returns Success or Running respectively. It returns Failure whenever all the children return Failure. When child $i$ returns Running or Success, the Fallback node stops sending ticks to the next child (if any) but keeps ticking all the children up to the child $i$.
\blue{
The Fallback node is graphically represented as} \BTNode{$?$} and its pseudocode is described in Algorithm~\ref{bts:alg:fallback}.
\begin{algorithm2e}[t]
\SetKwProg{Fn}{Function}{}{}

\Fn{Tick()}
{
  \For{$i \gets 1$ \KwSty{to} $N$}
  {
    \ArgSty{childStatus} $\gets$ \ArgSty{child($i$)}.\FuncSty{Tick()}\\
    \uIf{\ArgSty{childStatus} $=$ \ArgSty{Running}}
    {
      \Return{Running}
    }
    \ElseIf{\ArgSty{childStatus} $=$ \ArgSty{Success}}
    {
      \Return{Success}
    }
  }
  \Return{Failure}
  }
  \caption{Pseudocode of a Fallback operator with $N$ children}
    \label{bts:alg:fallback}
\end{algorithm2e}

\subsubsection*{Parallel}
When the Parallel node receives ticks, it routes them to all its children. It returns Success if a number $M$ of children return Success, it returns Failure if more than $N - M$ children return Failure, and it returns Running otherwise.
\blue{
The Parallel node is graphically represented as \BTNode{$\rightrightarrows$}} and its pseudocode is described in Algorithm~\ref{bts:alg:parallel}.

\begin{algorithm2e}[t]
\SetKwProg{Fn}{Function}{}{}

\Fn{Tick()}
{
  \ForAll{$i \gets 1$ \KwSty{to} $N$}
  {
    \ArgSty{childStatus}[i] $\gets$ \ArgSty{child($i$)}.\FuncSty{Tick()}\\
    }
    \uIf{$\Sigma_{i: \ArgSty{childStatus}[i]=Success} = M$}
    {
      \Return{Success}
    }
    \ElseIf{$\Sigma_{i: \ArgSty{childStatus}[i] =Failure} > N - M $}
    {
      \Return{Failure}
    
  }\Else{
  \Return{Running}
  }
  }
    \caption{Pseudocode of a Parallel operator with $N$ children}
  \label{bts:alg:parallel}
\end{algorithm2e}


\subsubsection*{Decorator}
A Decorator node represents a particular control flow node with only one child. When a Decorator node receives ticks, it routes them to its only child and it returns a status according to custom-made policy. BT designers use decorator nodes to introduce additional semantic or to change the return status of a node. Examples of decorators found in the literature are: the \emph{Inverter}, which changes the return status from Success to Failure and vice-versa, the \emph{Retry}, which re-sends ticks to a failed child while returning Running to its parent, and \emph{Timeline}, which sends ticks only for a limited time. It is 
 represented as a rhombus.

\subsubsection*{Action}
An action performs some operations as long as it receives ticks. It returns Success whenever the operations are completed and Failure if the operations cannot be completed. It returns Running otherwise. When a running Action no longer receives ticks, it aborts its execution. An Action node is graphically represented by a rectangle and its pseudocode is described in Algorithm~\ref{bts:alg:action}. The function call \emph{DoAPieceOfComputation$()$} may be executed in the BT itself or may represent a call to a service running on a separate component on the same or a remote machine. In these latter cases, a middleware usually handles the communication between the BT and the component. 
%

\begin{algorithm2e}[h]
\SetKwProg{Fn}{Function}{}{}

\Fn{Tick()}
{
 \ArgSty{DoAPieceOfComputation()} \\
    \uIf{action-succeeded}
    {
      \Return{Success}
    }
    \ElseIf{action-failed}
    {
      \Return{Failure}
    }
    \Else
    {
    \Return{Running}
    }
   }
  \caption{Pseudocode of a BT Action}
  \label{bts:alg:action}
\end{algorithm2e}


\subsubsection*{Condition}
Whenever a Condition node receives ticks, it checks if a proposition is satisfied or not. It returns Success or Failure accordingly. A Condition is graphically represented by an ellipse and its pseudocode is described in Algorithm~\ref{bts:alg:condition}.

\begin{algorithm2e}[h!]
\SetKwProg{Fn}{Function}{}{}

\Fn{Tick()}
{
    \uIf{condition-true}
    {
      \Return{Success}
    }
    \Else
    {
      \Return{Failure}
    }
   }
  \caption{Pseudocode of a BT Condition}
  \label{bts:alg:condition}
\end{algorithm2e}

\blue{

\subsection{Classical BTs semantic and other architectures}
\label{sec:btvsothers}
In the literature there is evidence that BTs generalize 
successful control architectures such as the 
 Subsumption architecture, Teleo-reactive Paradigm, Decision Tree and Finite State Machines (FSM)~\cite{BTBook}.

From a theoretical standpoint, every execution described by a BT can be characterized by a FSM and vice-versa. However, due to the number of transitions,
using a FSM as a control architecture leads to complex structures, as described in the literature~\cite{BTBook}.
Despite BTs generalize different control architectures, there are a few disadvantages on using the vanilla version of BTs, they comprises:
\begin{itemize}
\item Complexity of the BT engine. The BT engine's  implementation can become complex using single-threaded sequential programming as the actions may be executed asynchronously, as described in Section~\ref{sec:syncvsasyc}.
\item Checking all the conditions may be expensive. A BT needs to check several conditions to realize a task execution. In some applications, checks are too expensive or even unfeasible. In vanilla BTs implementations, presents more costs than advantages. For that, BTs have been extended with memory nodes, as described in Section~\ref{sec:memory}.
\item Sometimes a sequential (non reactive) execution is just fine. In applications where the robot
operates in a very structured environment that are predictable in space and time, BTs do
not have advantages over simpler architectures.
\item BT tools are less mature. Although there is software for developing BTs, it is still
far behind the amount and maturity of the software available developed for, e.g., FSMs. The BT community extended the BT's semantic by introducing node parameters and infra-node message passing as described in Section~\ref{sec:param}.
\end{itemize}

These disadvantages pushed the community to propose the extensions described in this letter.

}

\newpage
\section{Composition Nodes with Memory}
\label{sec:memory}
In this section, we describe the development of \emph{composition nodes with memory}. We provide their definition and pseudocode. \blue{They are also known as \emph{stateful} nodes~\cite{klockner2015behavior}. }

To provide for the reactivity in the BT's execution, the Sequence
and Fallback nodes keep sending ticks to the children to the left of a running child, to verify whether to re-execute a previous child and preempt the one that is currently running. However, there are cases in which the user knows that a child, once executed, does
not need to be re-evaluated (e.g., under the closed world assumption, there are no unexpected changes in the environment). 
To avoid
the unnecessary re-execution of some nodes, and save computational resources, the BT community proposed
 composition nodes with memory~\cite{millington2009artificial}.
A composition node with memory remembers the children that have been executed, preventing their useless re-evaluation.

\begin{figure}[h!]

        \begin{subfigure}[b]{0.3\columnwidth}
                \centering
                \includegraphics[width=\columnwidth]{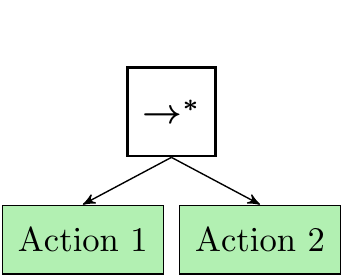}
                \caption{Sequence composition with memory.}
                \label{bts.fig.starnonreac}
        \end{subfigure}          
        ~ 
        \begin{subfigure}[b]{0.48\columnwidth}
                \centering
                \includegraphics[width=\columnwidth]{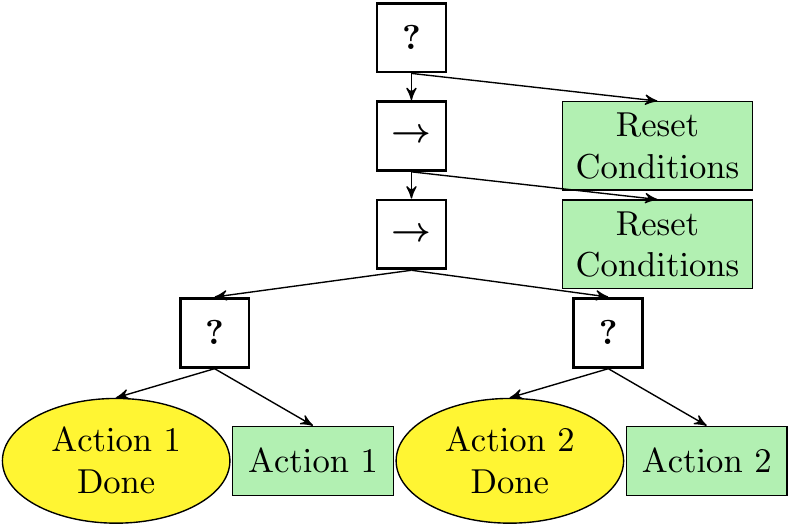}
                \caption{BT that emulates the execution of the Sequence composition with memory using classical nodes.}
                \label{bts.fig.starreac}              
        \end{subfigure}
        \caption{Relation between memory and memory-less BT nodes.}
                        \label{bts.fig.mem}              
\end{figure}

There exist different policies to reset the memory. We describe the most common policy, which consists of resetting the memory when the node returns either
Success or Failure, so that at the next activation, all children are re-considered for execution. Nodes with memory are graphically represented with the symbol \say{$*$} as superscript. Their execution can be obtained by
employing standard (memoryless) control flow nodes and adding auxiliary conditions and shared memory~\cite{BTBook}, as in Figure~\ref{bts.fig.mem}.

\subsubsection*{Sequence with memory}

The state of this node contains the index of the \emph{current} child. The node routes ticks to the current child. It returns Failure or Running whenever the current child returns Failure or Running, respectively. It returns Success when all the children returned Success. The node resets the index when a child returns Failure or when all the children returned Success. 
\blue{
A Sequence node with memory is graphically represented as \BTNode{$\rightarrow^*$} } and its pseudocode is described in Algorithm~\ref{bts:alg:sequencemem}.

\begin{algorithm2e}[h]
\SetKwProg{Fn}{Function}{}{}
\SetKwProg{Fn}{Function}{}{}
\Fn{Main}
{
\FuncSty{Reset()}
}
\Fn{Reset()}{
$i \gets 1$\\
}

\Fn{Tick()}
{
    \ArgSty{childStatus} $\gets$ \ArgSty{child($i$)}.\FuncSty{Tick()}\\
    \uIf{\ArgSty{childStatus} $=$ \ArgSty{Running}}
    {
      \Return{Running}
    }
    \ElseIf{\ArgSty{childStatus} $=$ \ArgSty{Failure}}
    {
    \FuncSty{Reset()}\\
      \Return{Failure}
    }
    \Else{
    	\uIf{$i<N$}
    	{
    	$i \gets i + 1$\\
    	}
    	\Else{
    	\FuncSty{Reset()}\\
      \Return{Success}
      }
    }
  }
  \caption{Pseudocode of a Sequence operator with memory with $N$ children}
  \label{bts:alg:sequencemem}
\end{algorithm2e}

\subsubsection*{Fallback with memory}

\blue{
A Fallback node with memory is graphically represented as \BTNode{$?^*$}} and its behavior is similar to the one of the Sequence node with memory, with the return status Failure in place of Success and vice-versa.

%
%
 
 
\subsubsection*{Parallel with memory}

When it receives ticks, it routes them to all the currently running children, stored in the memory. It returns Success if $M$ children return Success, it returns Failure if more than $N - M$ children return Failure, and it returns Running otherwise.
\blue{
The Parallel node with memory is graphically represented as
\BTNode{$\rightrightarrows^*$}} and its pseudocode is described in Algorithm~\ref{bts:alg:parallelmem}.

\begin{algorithm2e}[h]
\SetKwProg{Fn}{Function}{}{}
\Fn{Main}
{
\FuncSty{Reset()}
}
\Fn{Reset()}{
  \ForAll{$i \gets 1$ \KwSty{to} $N$}
  {
  \ArgSty{childDone}[i]$\gets$\ArgSty{False}
  }
  }
\Fn{Tick()}
{
  \ForAll{$i \gets 1$ \KwSty{to} $N$}
  {
  	\uIf{\ArgSty{childDone}[i]=\ArgSty{False}}
  	{
  	    \ArgSty{childStatus}[i] $\gets$ \ArgSty{child($i$)}.\FuncSty{Tick()}\\
  	      	\uIf{\ArgSty{childStatus}[i]!=Running}
					{\ArgSty{childDone}[i]$\gets$\ArgSty{True}}
  	}

    }
    \uIf{$\Sigma_{i: \ArgSty{childStatus}[i]=Success} = M$}
    {
        	\FuncSty{Reset()}\\
      \Return{Success}
    }
    \ElseIf{$\Sigma_{i: \ArgSty{childStatus}[i] =Failure} > N - M $}
    {
        	\FuncSty{Reset()}\\
      \Return{Failure}
    
  }\Else{
  \Return{Running}
  }
  }
    \caption{Pseudocode of a Parallel operator with memory  with $N$ children}
  \label{bts:alg:parallelmem}
\end{algorithm2e}

Some BT implementations,  do not include the \emph{Running} return status~\cite{millington2009artificial}. Instead, they let each Action run until it returns either \emph{Failure} or \emph{Success}. We denote these BTs as  \emph{non-reactive} since they do not activate actions other than those currently active and cannot react to changes. This is a relevant limitation of non-reactive BTs, which was noted by the authors of such libraries~\cite{millington2009artificial}. A non-reactive BT can be seen as a BT with only memory nodes. As reactivity is one of the key strengths of BTs, non-reactive BTs are rarely used in the literature. 
 
\section{Node parameters}
\label{sec:param}
In this section, we describe the development of \emph{nodes with parameters}. In the example presented in Section~\ref{sec:introduction}, we saw how some BTs require an implementation that allows node parametrization in terms of arguments passing and communication mechanisms between nodes.
Node parameters make BTs more flexible and reusable. In particular, the designer gains the ability to parametrize a particular task. Such parameters can either be \emph{explicit} in the form of arguments passed to the node (at design time) or \emph{implicit} in the form of a message passed to the node (at run time). 

\subsubsection*{Explicit arguments}

Leaf nodes describe a specific robot skill that may run differently according to the situation (e.g., the action \emph{Grasp} that can be performed with the \emph{left hand} or the \emph{right hand}, the condition \emph{Is robot in $<$room$>$}, etc.).
This supports the reuse of nodes similar to functions in computer languages or, more properly in our view, object-oriented programming. We can consider a leaf node as an object class with an internal state, in which member variables are assigned to the values of the argument passed from the constructor (we believe this analogy is more appropriate because a node is activated several times when it receives a tick from the BT engine and it has, therefore, an internal state that is updated every time the node receives a tick).

Some BT designers apply the argument assignments to an entire subtree with an exposed parameter interface~\cite{shoulson2011parameterizing,safronov2019asynchronous}. For subtrees, the interface inherits the parameter interface from its children. Such inheritance must avoid argument clash. From an object-oriented programming perspective, this is directly analogous to multiple inheritances. We can either implement the argument passing with the BT description sent to the engine or by defining a \say{configuration object} and which is passed by the BT engine to the subtree~\cite{gobehave}.

 
\subsubsection*{Implicit infra-node message passing}
The communication between BTs leaf nodes
traditionally relies on a \emph{blackboard}, a centralized shared memory of which all interested parties have access.
Some designers found the use of blackboards inconvenient for large BTs because it jeopardizes subtree reuse. Two encapsulated subtrees may both use
the same fields of the blackboard several layers down in their hierarchy, and could overwrite each other in a way that is difficult to trace\cite{tenorth2019controlling}. Moreover, some BT designers found it difficult to track the data stored in the blackboard~\cite{shoulson2011parameterizing}. Some BT implementations attempts to tame these problems by defining the \emph{scope} of some entries in blackboards~\cite{tenorth2019controlling, pytrees}. Another use of blackboards is for a BT to store the results of expensive computation, which takes place concurrently within the BT or in another branch of the BT composed with the parallel node to access the most recent result quickly. For example, an object detection algorithm may store all the objects detected in the current camera image in the blackboard. At the same time, a condition node checks whether a particular object visible or not.
Other solutions to the infra-node message passing rely on a middleware where the nodes share information using multi-variate dictionaries accessible via network APIs, as ROS Parameter Server~\cite{actionfw}.

\newpage
\section{Asynchronous Action Execution}
\label{sec:syncvsasyc}
In this section, we describe the \emph{asynchronous action execution} and the action's \emph{halting procedure},  crucial aspects to preempt and switch between actions. 

A BT orchestrates action and condition nodes by sending ticks to them and elaborating their return statuses. 
Action nodes control the robot, for example, by sending commands to the motors to perform a specific arm movement or to navigate to a given location. In a typical robot architecture, the actions are executed by independent components running on a distributed system. Therefore, the actions' execution gets delegated to different executables that communicate via a middleware (e.g., ROS).

An action node should return its status in a reasonably short time when ticked, so that tick propagation can proceed to guarantee reactivity. The propagation of the tick is blocking, and the time taken by nodes to return their status affects the step length of a \emph{safeguarding} BT~\cite{BTBook}. Thus, some BT designers split actions into several small steps; the action's execution starts when it receives its first tick, and it proceeds for another step only when the next tick is received. At the end of each step, a running action yields control back to the BT. This describes a \emph{synchronous execution}. However, with this solution, the tick frequency affects the speed at which an action progresses in time. Hence a designer must ensure sufficiently high tick frequency.

Other BT implementations execute actions \emph{asynchronously} with respect to the ticks. The action execution starts in a separated thread when it receives its first tick. Consequent ticks check if the action is still running or if the node has returned a Success or Failure status via an additional function call to get the node status. This solution results more intuitive in robotics applications because the designer of the leaf node has to write the code in a single block (without splitting it into several steps) and define only the conditions to return success or failure (see for example Algorithm~\ref{bts:alg:asyncaction}). However, the designer must ensure that an action's execution is halted safely whenever it no longer receives ticks.

\subsubsection*{Halting procedure}
Whenever the BT stops sending ticks to a running node to execute another one, this must suspend its execution. To ensure a stable switch,
the designer can either ensure that when the action returns its status, the robot lies in a stable state (in a synchronous execution setting) or explicitly define a specific function that ensures a safe halting routine (in an asynchronous execution setting), bringing the robot to a stable state (e.g., a bipedal robot places both feet on the ground when the action \emph{Walk} aborts). Such a function gets called whenever a BT no longer ticks a running node.

Looking back at the BT semantic presented in this letter, Algorithm~\ref{bts:alg:action} performs a step of computation at each tick, and the function \texttt{Tick()} runs in the same thread of the BT execution. Hence, the algorithm implements a synchronous execution of the actions. Algorithm~\ref{bts:alg:asyncaction} shows an implementation example of an asynchronous action.  The function \texttt{Tick()} runs in a separate thread of the BT execution. The function initializes the status as \emph{Running}, then it executes the code of the action, which will eventually change the status to  \emph{Success} or \emph{Failure}. The action implements also the function \texttt{Status()} that returns its internal status without executing the node.
Some BT implementations~\cite{BTCpp} of asynchronous nodes include the return status \emph{Idle} to indicate that the node was never executed or that the parent node read the return status Success or Failure. 

With asynchronous actions, the BT implementation needs to ensure that the thread executing the function  \texttt{Tick()} gets aborted correctly whenever the BT no longer sends ticks to a running node.
%
%
The implementation of composition nodes must also be adapted accordingly, calling the function \texttt{Tick()} of a child only when the latter is not running and the function \texttt{Halt()} of a node that is no longer receiving ticks. 

\begin{algorithm2e}[h]
\SetKwProg{Fn}{Function}{}{}

\Fn{Tick()}
{
 \ArgSty{current-status} $\gets$ \ArgSty{Running}\\
  \FuncSty{Action Implementation} \\
   }
   
   \Fn{Halt()}
{
  \FuncSty{Safe Abort Routine} \\

   }
  	 
   \Fn{Status()}
{
 	\Return{current-status}
   }
  \caption{Pseudocode of an Asynchronous BT Action}
  \label{bts:alg:asyncaction}

\end{algorithm2e}
A controversial aspect concerns the blocking nature of the function \texttt{Halt()}, i.e. if the flow control must be passed to the BT only after the function \texttt{Halt()} terminates. Given the nature of the activities performed by \texttt{Halt()} existing libraries implement it in a blocking fashion~\cite{BTCpp,gobehave}.

Other solutions do not include the function \texttt{Halt()}; instead, they implement the action's abortion procedures as separate actions to keep the BT semantic as simple as possible~\cite{paxton2017costar, pytrees, braintree}. This, however, may negatively impact the modularity and reusability of nodes within the BT.

\subsubsection*{Coroutines}
The library BehaviorTree.CPP~\cite{BTCpp} introduced \emph{CoroActionNodes} to execute actions. 
Instead of executing the function \texttt{Tick()} in a separate thread explicitly, it uses coroutines, a control structure where the flow control gets cooperatively passed between the two different routines (i.e., the tick traversal and the action node execution) until either the parent node no longer ticks the action or the action terminates its execution. With CoroActionNodes, the designer explicitly calls the function \texttt{setStatusRunningAndYield()}, whenever the BT can suspend the action execution and pass the flow control back to the parent node.
CoroActionNodes also implement the function \texttt{Halt()} to safely abort the execution. 
The use of coroutines for implementing BT actions results conveniently from a design and implementation point of view: We can implement actions as a single block of code (as for asynchronous actions) and explicitly define when the action execution can be suspended (as for synchronous actions). Unfortunately, it is a feature we found only in the library BehaviorTree.CPP.

%
%
%
%

\newpage

\newpage

\section{Available Open-source Libraries}
\label{sec:libraries}

\blue{The literature presents a complete review of BTs libraries~\cite{ghzouli2020behavior}, in contrast, in this section, we discuss how the most used libraries implement the features described above.\footnote{We searched for the  most active and documented BT repositories that are relevant for robotic applications.}

}
\blue{
 Table~\ref{tab.libs} synthesizes a comparison of the six most used open-source libraries. Most of the the libraries support memory nodes; in some cases, they consider memory nodes as \say{standard} nodes \cite{BTCpp,paxton2017costar}, and in other cases they are the only ones supported. The blackboard represent the common implementation of infra-nodes message passing. The asynchronous action execution, which is fundamental for BT implementation in robotics, appears to be surprisingly missing in some of them.
Moreover, a few of libraries allow the creation of BTs using a GUI in a drag-and-drop fashion and provide traces of execution for debugging purposes.}
 
 We now describe in detail the three most used libraries. 




\subsubsection*{BehaviorTree.CPP (BT.cpp)~\cite{BTCpp}} 
 It is a C++ library that allows both the definition of classical composition nodes and memory counterparts. A user can define an interface for the leaf nodes in terms of input and output \emph{ports}. With input ports, a designer can define arguments, as strings that can be parsed by the node, or keys to the blackboard's entry. An output port points to the blackboard's entry, whose value can be set inside the node. The library provides two ways of implementing asynchronous action by either executing the function \texttt{Tick()} in a separate thread or using \emph{coroutines} that uses only one thread.
Users can also define the BT by manually editing an eXtensible Markup Language (XML) file or through a GUI, available on a separate repository.

\subsubsection*{py\_trees~\cite{pytrees}}

\blue{It is a Python library that, to date,
does not contain the sequence node but only its memory counterpart but it contains the fallback node.}
The library does not allow argument passing to the node, and it implements only synchronous actions. The user must define the read/write access permission to keys on a blackboard for each leaf node.
The library provides run-time BT visualization in the following formats Unicode and Graphviz DOT.
%

\blue{\subsubsection*{CoSTAR~\cite{paxton2017costar}}
It is an end-user interface for authoring robot task plans that includes a BT-based user interface. It is implemented as a C++ library.
It allows the definition of classical composition nodes with memory only. The user can design the BT using a GUI. The library allows arguments to leaf nodes 
and it contains a blackboard.}
%
%
%


%
\begin{table}[h!]
\centering
\begin{tabular}{|c c c c c c c|} 
 \hline
 Name & Rea & Mem & Args & BB & Asyn  & GUI\\ [0.5ex] 
 \hline
	BT.cpp~\cite{BTCpp} & \cmark & \cmark & \cmark & \cmark & \cmark & \cmark \\ 
 \hline
 	py\_trees~\cite{pytrees} & \xmark & \cmark & \xmark & \cmark & \xmark & \xmark \\ 
 \hline
  	CoSTAR~\cite{paxton2017costar} & \xmark & \cmark & \cmark & \cmark & \xmark & \cmark \\ 
 \hline

  	BrainTree~\cite{braintree}  & \cmark & \cmark & \cmark & \cmark & \xmark & \xmark \\ 
 \hline
 
   go-behave~\cite{gobehave}  & \xmark & \cmark & \cmark & \cmark & \xmark & \xmark \\ 
 \hline
     	ActionFW~\cite{actionfw} & \cmark & \cmark & \xmark & \xmark & \cmark & \xmark \\\hline
\end{tabular}
\caption{Comparison in terms of \textbf{Rea}ctiveness, \textbf{Mem}ory nodes, \textbf{Arg}uments, \textbf{B}lack\textbf{B}oard, \textbf{Async}hronous nodes, and \textbf{G}\textbf{U}\textbf{I}.}
\label{tab.libs}
\end{table}

\newpage

\section{Behavior Trees in a Robotic Software Architecture}
\label{sec:example}

\blue{In this section, we suggest where BTs fit inside a robotic software architecture and we provide a use case example. We also provide a simplified example, available online.\footnote{\url{github.com/hsp-iit/behavior-stack-example}}}

\subsection{Abstraction Layers}
The robotic software is often partitioned into different layers of abstractions, each addressing different concerns; the layers' separation allows level-specific efficient solutions~\cite{ahmad2016software}. The number and separation of the layers are the subjects of controversial discussions; in this letter, we use the abstraction layers defined by the RobMoSys Robotic Software Component\footnote{\url{robmosys.eu/wiki/start}}, which we used in different projects.\footnote{\url{carve-robmosys.github.io}}$^,$\footnote{\url{scope-robmosys.github.io}} 
Following the abstraction above, we categorize the robotic software in:

 \subsubsection*{Mission Layer}
The layer where we design the higher-level goal for the robot to achieve (e.g., serve a customer). It may employ symbolic task planners, constraint solvers, and analysis tools to define a goal. It can be implemented as an algorithm~\cite{colledanchise2019towards, tadewos2019automatic}
or a human interface~\cite{coronado2018development,paxton2017costar,shepherd2018engineering} that synthesizes a BT. 
 \subsubsection*{Task Layer}
The layer where we design how the robot accomplishes a goal, disregarding the implementation details. In our context, it describes the BT.
 \subsubsection*{Skill Layer}
The layer where we design the basic capabilities (\emph{skills}) of a robot (e.g., grasp an object, go to a given location in the map, etc.). A skill describes the implementation of a leaf node of a BT. Each skill is implemented by orchestrating a set of services from the \emph{Service Layer}, orchestration involves retrieving or setting parameters, starting, and stopping services. A state machine well describes these operations. Concerning the BT engine, a skill can run in two ways:

- In the same executable, where the designer writes the source code of the skill inside a leaf node in the BT engine.

- In a separate executable, where the source code of the skill is written in a separate component that exposes the interface for calling
the functions \texttt{Tick()} and (possibly) the \texttt{Halt()}. A leaf node of the BT engine forwards the calls of the functions \texttt{Tick()} and \texttt{Halt()} to the corresponding component, using inter-process communication facilities made available by the Operating System or a middleware (e.g., through shared memory or the network).

\subsubsection*{Service Layer} The layer that contains entities that serve as the access point for the skills to command the robot. It describes the server side of a service for basic capabilities (e.g., getting the battery level, moving the base, etc.).

The other RobMoSys abstraction layers are: Function, Operating System, and Hardware. They represent software components that the robotic designer rarely develops, and therefore they fall beyond the scope of this letter.

%
%

\section{Use Case Example}
\label{sec:use}
\blue{
In this section, we provide a use case example to show how to build a full working robot behavior execution. An R1 robot has to fetch an empty bottle from a customer's hand and then bring it to a counter. The robot has to move the arm in a pre-grasp position and then close the hand to fetch the object. The BT in Figure~\ref{fig:ex:BTandFSM:BT} encodes such behavior. \blue{A video of the execution is available online.\footnote{The link will be available at the publication stage. Please watch the attached video.}}

}

In the example, we manually designed the BT using a GUI\footnote{\url{github.com/BehaviorTree/Groot}} (Mission Layer) and execute it via the BehaviorTree.CPP engine~\cite{BTCpp} (Task Layer). We run the skills in a separate executable than the one of the BT (Skill Layer).
We use pre-existing navigation	 and manipulation servers to interact with the robot's hardware (Service Layer).

We implement the skills as external executables as either YARP RFModule\footnote{The RFModule in YARP is an abstraction for components, it implements basic functions for handling the lifecycle of a modules, its parameters and it be extended with functions that are specific to each component.}~\cite{paul2014middle} or State Charts (SCs) defined via the SCXML formalism and executed via the QtSCXML engine\footnote{\url{doc.qt.io/qt-5/qtscxml-overview.html}}.
In either case, the skill exposes the interface to start or abort the skill execution.
Each state in a SC contains \texttt{C++11} source code. The SC may contain two states representing a succeeded and failed execution respectively.\footnote{Actions may not have a termination condition. E.g. move forwards.}
We found it convenient to use SCs for actions from a reusable perspective. Still, any other modeling framework for skills can fit our purpose, provided that it can handle the requests from the BT in terms of execution and abortion.  
The executable of a SC skill has two threads: one that handles the requests from the leaf node and executes the SC engine.

\blue{We implemented the action nodes a Coroutines (see Section~\ref{sec:syncvsasyc}), these actions}
forward the \emph{start} and \emph{stop} (for actions only) via the YARP middleware, respectively, when they receive the first tick and when they receive a halt. The details of the corresponding skill's executable are passed to the leaf node as an argument. Whenever a leaf node receives a tick, it sends the request to the corresponding external executable to start the skill; when the leaf node receives a halt, it sends the request to the corresponding external executable to stop the skill. We implemented actions as CoroActionNodes (see Section~\ref{sec:syncvsasyc}).  
\blue{The leaf node has parameters to set the name of the corresponding skill and communicate with each other by using a blackboard (see Section~\ref{sec:param}).}
\blue{We chose to not use nodes with memory, as any sequence of actions can be modeled as a SC inside a single BT action (see Figure~\ref{fig:ex:BTandFSM:action})}
We describe the skills interfaces using the YARP Thrift IDL~\cite{paul2014middle} and implemented them as the following Remote Procedure Calls (RPCs): \texttt{start()} that starts the execution of the skill and returns a boolean that describes if the skill was performed correctly or not; \texttt{stop()} that aborts the execution of the skill and returns void. 

\subsection{Skills Implementation Examples}


%
%

\begin{figure}[t]
\centering
\begin{subfigure}[t]{0.435\columnwidth}
\centering
\includegraphics[width=0.85\columnwidth]{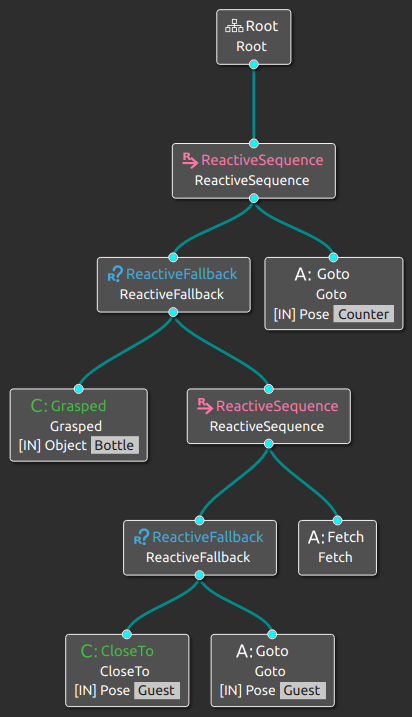}
\caption{BT.}
\label{fig:ex:BTandFSM:BT}

\end{subfigure}
\begin{subfigure}[t]{0.5\columnwidth}
\centering
\includegraphics[width=0.85\columnwidth]{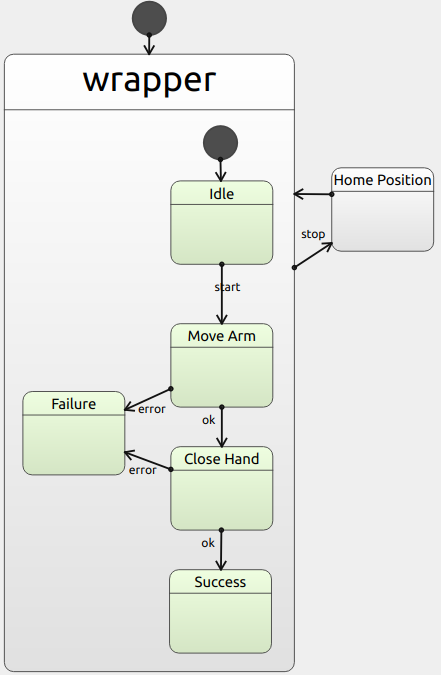}
\caption{Skill \say{Fetch} implemented as SC.}
\label{fig:ex:BTandFSM:action}
\end{subfigure}
 \vskip1.4em
\begin{subfigure}[b]{\columnwidth}
\fbox{\includegraphics[width=\columnwidth]{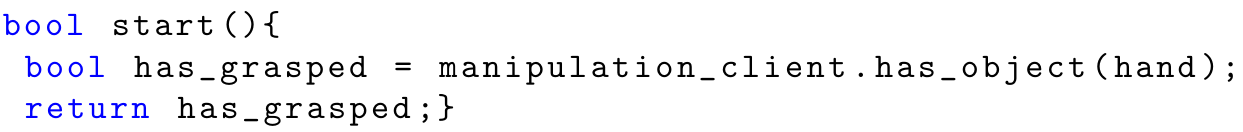}}
\caption{Function \texttt{start()} of the skill \say{Object Grasped} implemented as RFModule.}
\label{fig:ex:BTandFSM:condition}
\end{subfigure}
 \vskip1.4em
\caption{BT and skills of the use case.}
\label{fig:ex:BTandFSM}
\end{figure}

We developed the following skills:
\begin{itemize}
\item \emph{Object Grasped} that sends requests to a robot interface to check if the robot is holding an object. The Skill has one argument, \emph{hand}, that represents which hand to check.
Figure~\ref{fig:ex:BTandFSM:condition} shows a portion of its implementation as YARP RFModule.
\item \emph{Close to Pose} that sends requests to a localization server to check if the robot is close to a given pose. The skill has two arguments \emph{position} and \emph{threshold}.
\item \emph{Goto Pose} whose function \texttt{start()} sends requests to a navigation server (on the Service Layer) to compute a path and follow it with the robot's base. The function returns \texttt{False} if the navigation server cannot find a path to the destination. It returns \texttt{True} if the robot reaches the pose.
 The function \texttt{stop()} aborts the navigation when the BT switches execution from the Goto action nodes to other nodes.
\item \emph{Fetch} whose function \texttt{start()} starts a SC (see Figure~\ref{fig:ex:BTandFSM:action}). The SC sends requests to a manipulation server (on the Service Layer) to move the arm in a pre-grasp position and then to close the hand.
This function returns \texttt{False} if the SC goes in the \emph{failure} state, it returns \texttt{True} if it goes in the \emph{success} state.
The \texttt{stop()} triggers the transition \emph{stop} in the SC, making the arm move to a home position if nothing is grasped. 
\end{itemize}

\blue{
To summarize the example above uses: no memory nodes, as the linear sequence of actions are encoded as SC; action execution with Coroutines; external executables that forward commands to the robot; infranode message passing implemented as a blackboard; and nodes with parameters.
}

\section{Concluding Remarks}
\label{sec:conclusions}

In this letter, we outlined the practical aspects of BTs in robotics and outlined state-of-the-art solutions to date. 
\blue{The analysis highlighted that a designer, to take advantage of BTs fully, relies on a execution engine with specific characteristics.
Some of these characteristics are still debated, and different existing libraries implement different solutions, such as handling a safe preemption, the asynchronous action execution, and the shared memory.
We also showed  BTs fit in generic robotic software architectures and presented a practical use case example.}
We believe that this letter provides useful information for both the BT designers that want to fully understand the characteristics and limitations of a particular BT execution engine and the software developers that want to implement their own BT execution engine.


\bibliographystyle{IEEEtran}
\bibliography{refs}
\end{document}